\documentclass{article} 
\usepackage{iclr2023,times}

\usepackage[utf8]{inputenc} 
\usepackage[T1]{fontenc}    
\usepackage[hidelinks]{hyperref}       
\usepackage{url}            
\usepackage{booktabs}       
\usepackage{amsfonts}       
\usepackage{nicefrac}       
\usepackage{microtype}      
\usepackage{xcolor}         
\usepackage{graphicx}
\usepackage{multirow}
\usepackage{tabularx}
\usepackage{arydshln}
\usepackage[T5]{fontenc}
\iclrfinaltrue

\title{PhoGPT: \\ Generative Pre-training for Vietnamese}

\author{Dat Quoc Nguyen, Linh The Nguyen, Chi Tran, Dung Ngoc Nguyen, Dinh Phung, Hung Bui  \\
VinAI Research, Hanoi, Vietnam\\
  \small{\texttt{\{v.datnq9, v.linhnt140, v.chitb, v.dungnn28, v.dinhpq2, v.hungbh1\}@vinai.io}}
}

\begin{document}

\maketitle

\begin{abstract}
We open-source a state-of-the-art 4B-parameter generative model series for Vietnamese, which includes the base pre-trained monolingual model PhoGPT-4B and its chat variant, PhoGPT-4B-Chat. The base model, PhoGPT-4B, with exactly 3.7B parameters, is pre-trained from scratch on a Vietnamese corpus of 102B tokens, with an 8192 context length, employing a vocabulary of 20480 token types. The chat variant, PhoGPT-4B-Chat, is the modeling output obtained by fine-tuning PhoGPT-4B on a dataset of 70K instructional prompts and their responses, along with an additional 290K conversations. In addition, we also demonstrate its superior performance compared to previous open-source models. 
Our PhoGPT models are available at: \url{https://github.com/VinAIResearch/PhoGPT}
\end{abstract}

\section{Introduction}

Undoubtedly, the success of large language models (LLMs), particularly decoder-only transformer-based generative models such as ChatGPT/GPT-4, 
LLaMA/LLaMA-2 \citep{touvron2023llama,touvron2023llama2}, Mistral \citep{jiang2023mistral} and Falcon \citep{falcon40b},  stands out as one of the most significant achievements in recent AI research and development. However, that success has largely been limited to English. 

For Vietnamese, we now release the 4B-parameter base pre-trained monolingual model, PhoGPT-4B, along with its chat variant, PhoGPT-4B-Chat, as open-source. We pre-train the base model PhoGPT-4B from scratch on a Vietnamese corpus of 102B tokens for two epochs, with an 8192 context length. Here, it employs a Vietnamese-specific byte-level BPE tokenizer with a vocabulary of 20480 tokens. We further fine-tune the base model on a dataset of 70K instructional prompts and their responses, along with an additional 290K conversations, resulting in the chat variant, PhoGPT-4B-Chat. We demonstrate its strong performance compared to previous closed-source and open-source 7B-parameter models. 

Our goal is to provide comprehensive and powerful LLMs for Vietnamese, facilitating future research and applications in generative Vietnamese NLP. Our PhoGPT can be used with popular libraries such as ``transformers'' \citep{wolf-etal-2020-transformers}, ``vllm'' \citep{kwon2023efficient} and ``llama.cpp''.\footnote{\url{https://github.com/ggerganov/llama.cpp}}




\section{PhoGPT}

\subsection{PhoGPT-4B: Model architecture and Pre-training}

PhoGPT-4B is a Transformer decoder-based model \citep{NEURIPS2020_1457c0d6,NIPS2017_7181},  which incorporates (Triton) flash attention \citep{dao2022flashattention} and ALiBi \citep{press2022train} for context length extrapolation. We train a Vietnamese-specific byte-level BPE tokenizer with a vocabulary of 20480 tokens using the ``tokenizers'' library.\footnote{\url{https://github.com/huggingface/tokenizers}}  In addition, we use a ``max\_seq\_len'' of 8192, ``d\_model'' of 3072, ``n\_heads'' of 24 and ``n\_layers'' of 32, resulting in a model size of 3.7B  parameters ($\sim$4B).  Utilizing the Mosaicml ``llm-foundry'' library \citep{MosaicML2023Introducing},\footnote{\url{https://github.com/mosaicml/llm-foundry}} 
we pre-train PhoGPT-4B from scratch  on a 482GB deduplicated and cleaned pre-training corpus of Vietnamese texts ($\sim$102B tokens) for {two epochs}. Our pre-training Vietnamese corpus consists of:\footnote{Last crawling/cutoff date: 31/05/2023.}

\begin{itemize}
\item 1GB of Wikipedia texts (version 20/05/2023);
\item 1.5GB of medical-related texts crawled from a wide range of publicly available and medical domain-specific websites such as medical journals and universities;\footnote{The content extracted from these sources contains no private data about the patients.}
\item 3GB of publicly available books spanning a range of genres;
\item 12GB of legal data crawled from \url{thuvienphapluat.vn}
 and \url{lawnet.vn};
\item a 40GB variant of the "binhvq" news corpus (version 21/05/2021);\footnote{\url{https://github.com/binhvq/news-corpus}}
\item an 88GB variant of the Vietnamese OSCAR-2301 subset;\footnote{\url{https://huggingface.co/datasets/oscar-corpus/OSCAR-2301}}
\item a 336GB variant of the Vietnamese mC4 subset.\footnote{\url{https://huggingface.co/datasets/allenai/c4/tree/mC4_3.1.0}}
\end{itemize}

\subsection{PhoGPT-4B-Chat: Supervised fine-tuning}

We then fine-tune the base pre-trained PhoGPT-4B using a dataset consisting of 70K instructional prompts and their responses, along with an additional 290K conversations, constructed by concatenating the following sources:

\begin{itemize}
\item 500 instructional prompt and response pairs for poem writing, 500 for essay writing, 500 for spelling correction,  500 for single-document summarization and 1000  for context-based question answering;
\item 67K instructional prompt and response pairs from the Vietnamese subset of Bactrian-X \citep{li2023bactrianx};
\item 20K Vietnamese-translated ChatAlpaca  conversations;\footnote{\url{https://github.com/cascip/ChatAlpaca}}
\item 40K Vietnamese-translated ShareGPT conversations (without code and mathematics);\footnote{\url{https://huggingface.co/datasets/anon8231489123/ShareGPT_Vicuna_unfiltered}} 
\item 230K Vietnamese-translated UltraChat conversations \citep{ding2023enhancing};\footnote{\url{https://huggingface.co/datasets/HuggingFaceH4/ultrachat_200k} (including both train \& test sets)}
\end{itemize}
 
The resulting fine-tuned model is named PhoGPT-4B-Chat.

\section{Evaluation}


We compare PhoGPT-4B-Chat with the closed-source models GPT-4-0125-preview, GPT-3.5-turbo and Gemini Pro 1.0, as well as other open-source models, including:

\begin{itemize}
\item Vistral-7B-Chat is the modeling output obtained by continually pre-training Mistral-7B \citep{jiang2023mistral} on a diverse corpus of Vietnamese texts and then performing supervised fine-tuning using a diverse instructional and conversational dataset.\footnote{\url{https://huggingface.co/Viet-Mistral/Vistral-7B-Chat}}

\item SeaLLM-7B-v2 is the modeling output obtained by continually pre-training Mistral-7B on a multilingual corpus from Southeast Asian (SEA) languages, including Vietnamese, and then performing supervised fine-tuning using instructional question and answer pairs.\footnote{\url{https://huggingface.co/SeaLLMs/SeaLLM-7B-v2}}

\item Sailor-7B-Chat\footnote{\url{https://huggingface.co/sail/Sailor-7B-Chat}} and Sailor-4B-Chat \footnote{\url{https://huggingface.co/sail/Sailor-4B-Chat}} are the modeling outputs obtained by continually pre-training Qwen1.5-7B and Qwen1.5-4B \citep{bai2023qwen} on 400B tokens from SEA languages, including Vietnamese, and then performing supervised fine-tuning using instructional question and answer pairs.

\item VBD-LLaMA2-7B-50b-Chat is the modeling output obtained by continually pre-training LLaMA-2-7B \citep{touvron2023llama2} on a data combination of 40B Vietnamese tokens and 16B English tokens and then performing supervised fine-tuning using 2M instructional and conversational samples.\footnote{\url{https://huggingface.co/LR-AI-Labs/vbd-llama2-7B-50b-chat}}

\item VinaLLaMA-7B-Chat is the modeling output obtained by continually pre-training LLaMA-2-7B on a data combination of 100B English tokens, 230B Vietnamese tokens (from books and news), and 500B automatically generated Vietnamese tokens, and then performing supervised fine-tuning using 1M instructional and conversational samples.\footnote{\url{https://huggingface.co/vilm/vinallama-7b-chat}}

\item Gemma-7B-it is the 7B instruct version of the Gemma model \citep{gemmateam2024gemma}.\footnote{\url{https://huggingface.co/google/gemma-7b-it}}

\end{itemize}

\begin{table*}[!t]
\centering
\begin{tabular}{l|c|c}
\hline 
\textbf{Model} & \textbf{All truthful questions} & \textbf{Vietnam-specific} \\
\hline
PhoGPT-4B-Chat	& \underline{41.7	(83 / 199)}	& \textbf{43.5	(64 / 147)} \\
GPT-4-0125-preview		& \textbf{44.7	(89 / 199)}		& 39.5	(58 / 147)  \\
GPT-3.5-turbo		& 29.1	(58 / 199)		& 22.4	(33 / 147)  \\
Gemini Pro 1.0		& 39.7	(79 / 199)		& 34.7	(51 / 147)  \\
Vistral-7B-Chat		& 41.2	(82 / 199)		& \underline{42.9	(63 / 147)}  \\
Sailor-7B-Chat		& 28.6	(57 / 199)		& 27.9	(41 / 147)  \\
Sailor-4B-Chat		& 15.6	(31 / 199)		& 14.3	(21 / 147)  \\
SeaLLM-7B-v2		& 20.6	(41 / 199)		& 13.6	(20 / 147)  \\
VBD-Llama2-7B-50B-Chat		& 15.6	(31 / 199)		& 10.9	(16 / 147)  \\
Vinallama-7B-Chat		& 11.1	(22 / 199)		& 8.2	(12 / 147)  \\
Gemma-7B-it		& 8.0	(16 / 199)		& 6.1	(9 / 147)  \\
\hline 
\end{tabular}
\caption{Obtained results.}
\label{table:results}
\end{table*}


Our empirical study employs the Vietnamese truthful question-answering dataset ViTruthfulQA \citep{nguyen-etal-2023-vigptqa}, comprising 199 questions.\footnote{The original dataset consists of 213 questions; however, 14 of them are not questions or are unclear, e.g. ``Tên gọi nào không phải là một loại trái cây Việt Nam?'' (Which name is not a Vietnamese fruit?). Therefore, we remove them, resulting in a final evaluation set of 199 questions.} Each of the 199 questions is fed into 11 experimental models to generate responses, which are then anonymously shuffled. Here, we utilize the greedy search decoding method, which is more suitable for LLM comparison \citep{lin-chen-2023-llm}. Two annotators independently assess each generated response on whether the response is correct or not. A response is annotated as ``correct'' only if it contains the accurate answer for the corresponding question without any hallucinated information. We host a discussion session with the annotators to resolve annotation conflicts. 


Table \ref{table:results} presents the accuracy results obtained, showing that overall, our 4B-parameter PhoGPT-4B-Chat is highly competitive compared to the closed-source GPT-4 model, yielding better accuracy scores than the remaining closed-source models GPT-3.5-turbo and Gemini Pro 1.0, as well as all open-source baselines. 
Furthermore, when it comes to 147 out of the 199 questions that specifically ask for information related to Vietnam, PhoGPT-4B-Chat achieves the highest accuracy. 

\section{Conclusion}

We have introduced state-of-the-art open-source  4B-parameter LLMs for Vietnamese, including the base pre-trained PhoGPT-4B and its chat variant, PhoGPT-4B-Chat. 
We hope that these models will foster future research and applications of Vietnamese LLMs.

\section*{Limitations}

PhoGPT has certain limitations. For example, it is not good at tasks involving reasoning, coding or mathematics. PhoGPT may generate harmful, hate speech, biased responses, or answer unsafe questions. Users should be cautious when interacting with PhoGPT that can produce factually incorrect output.

\section*{Acknowledgments}

We extend our thanks to Nhung Nguyen (v.nhungnt89@vinai.io) for crawling and pre-processing health data and to Thien Huu Nguyen (v.thiennh4@vinai.io) for the initial discussions.

\bibliographystyle{iclr2023_conference}
\bibliography{refs}

\end{document}